\documentclass[sigconf, anonymous=False]{acmart}
\AtBeginDocument{%
  \providecommand\BibTeX{{%
    \normalfont B\kern-0.5em{\scshape i\kern-0.25em b}\kern-0.8em\TeX}}}

\usepackage{multirow}
\usepackage{float} 
\usepackage{amsmath}
\usepackage[ruled,linesnumbered]{algorithm2e}
\usepackage{graphicx}
\usepackage{subfigure}
\usepackage{epstopdf}
\usepackage{tablefootnote}
\usepackage[flushleft]{threeparttable}
\usepackage{diagbox}
\usepackage{booktabs}
\usepackage{siunitx}
\usepackage{footmisc}
\usepackage{footnote}
\usepackage{enumitem}
\usepackage[normalem]{ulem}
\useunder{\uline}{\ul}{}

\begin{document}

\title{LeCaRDv2: A Large-Scale Chinese Legal Case Retrieval Dataset}
\author{Haitao Li}
\affiliation{DCST, Tsinghua University}
\affiliation{Quan Cheng Laboratory}
\affiliation{Beijing 100084, China}
\email{liht22@mails.tsinghua.edu.cn}

\author{Yunqiu Shao}
\affiliation{DCST, Tsinghua University}
\affiliation{Quan Cheng Laboratory}
\affiliation{Beijing 100084, China}
\email{shaoyq18@mails.tsinghua.edu.cn}

\author{Yueyue Wu}
\affiliation{DCST, Tsinghua University}
\affiliation{Quan Cheng Laboratory}
\affiliation{Beijing 100084, China}
\email{wuyueyue@mail.tsinghua.edu.cn}

\author{Qingyao Ai}
\affiliation{DCST, Tsinghua University}
\affiliation{Quan Cheng Laboratory}
\affiliation{Beijing 100084, China}
\email{aiqy@tsinghua.edu.cn}

\author{Yixiao Ma}
\affiliation{DCST, Tsinghua University}
\affiliation{Quan Cheng Laboratory}
\affiliation{Beijing 100084, China}
\email{mayx20@mails.tsinghua.edu.cn}

\author{Yiqun Liu}
\authornote{Corresponding author}
\affiliation{DCST, Tsinghua University}
\affiliation{Quan Cheng Laboratory}
\affiliation{Beijing 100084, China}
\email{yiqunliu@tsinghua.edu.cn}


\begin{abstract}
As an important component of intelligent legal systems, legal case retrieval plays a critical role in ensuring judicial justice and fairness. However, the development of legal case retrieval technologies in the Chinese legal system is restricted by three problems in existing datasets: limited data size, narrow definitions of legal relevance, and naive candidate pooling strategies used in data sampling.

To alleviate these issues, we introduce LeCaRDv2, a large-scale \textbf{Le}gal \textbf{Ca}se \textbf{R}etrieval \textbf{D}ataset (version 2). It consists of 800 queries and 55,192 candidates extracted from 4.3 million criminal case documents. To the best of our knowledge, LeCaRDv2 is one of the largest Chinese legal case retrieval datasets, providing extensive coverage of criminal charges. Additionally, we enrich the existing relevance criteria by considering three key aspects: characterization, penalty, procedure. This comprehensive criteria enriches the dataset and may provides a more holistic perspective. Furthermore, we propose a two-level candidate set pooling strategy that effectively identify potential candidates for each query case. 
It's important to note that all cases in the dataset have been annotated by multiple legal experts specializing in criminal law. Their expertise ensures the accuracy and reliability of the annotations. We evaluate several state-of-the-art retrieval models at LeCaRDv2, demonstrating that there is still significant room for improvement in legal case retrieval. The details of LeCaRDv2 can be found at the anonymous website https://github.com/THUIR/LeCaRDv2.

\end{abstract}


\maketitle

\section{Introduction}
As a fundamental component of intelligent legal systems, legal case retrieval technology plays an essential role in ensuring justice in judgments. In countries with case law system, judges need to make a final decision based on the previous judgments of relevant cases~\cite{shulayeva2017recognizing}. In countries with statutory law system, when a case is presented to the court, extensive relevant cases are reviewed to avoid inappropriate judgments~\cite{hamann2019german}. With the rapid growth of digitized legal cases, more and more researchers have started to look into the problem and try to apply natural language processing (NLP) and information retrieval (IR) techniques to address the problem of legal case retrieval~\cite{shao2020bert,xiao2021lawformer,chalkidis2020legal,li2023constructing}.

To facilitate relevant research, researchers have begun to build human-annotated datasets for legal case retrieval in recent years ~\cite{rabelo2022overview,rabelo2021coliee,ma2021lecard,locke2018test}. For instance, Juliano Rabelo et al. ~\cite{rabelo2022overview,rabelo2021coliee} provide the Canadian legal dataset COLIEE, which belongs to the case law system. Additionally, Ma et al. ~\cite{ma2021lecard} introduce new relevance judgment criteria for the Chinese statutory law system and construct LeCaRDv1, a Chinese legal case retrieval dataset comprising queries and corresponding relevant case documents. Despite its valuable contributions to the advancement of legal case retrieval techniques in the Chinese legal system, there are still three primary challenges remaining unsolved:

\begin{itemize}[leftmargin=*]
    \item \textbf{Limited Data.}
    The LeCaRDv1 only contains a hundred queries with a limited number of annotated case documents, which may not be sufficient for the training of large language models and providing reliable evaluation results. More specifically, LeCaRDv1 has 10,700 candidate cases and 107 query cases covering 20 charges. The small query set size and charge coverage rate could limit the scope of the application of the dataset. 
   
    \item \textbf{Narrow Definition of Legal Relevance.}
    The relevance criterion of LeCaRDv1 focuses only on the fact description section of a case and ignores the similarities in penalty and procedure. When it comes to creating high-quality datasets, relevance criterion is a fundamental concern especially in the legal field.
    In general, the relevance in the legal field differs from generic textual similarity and goes beyond topic relevance~\cite{van2017concept}.
    In case law systems datasets like COLIEE, a relevant case is typically defined as a previous case cited by the query case. This criterion may not be applicable in countries with statutory law systems, such as China. As a pioneer, LeCaRDv1 proposes new criteria for guiding experts to determine relevance based on critical factors. However, it only concentrates on the fact section, potentially resulting in partial understanding and biased annotations of result relevance.

    \item \textbf{Naive Candidate Pooling Strategy.}
    LeCaRDv1 employs three retrieval models, namely TF-IDF~\cite{TF-IDF}, BM25~\cite{robertson2009probabilistic}, and LMIR~\cite{LMIR}, to construct a 100-case pool for each query. However, these methods primarily rely on lexical matching and exhibit similar characteristics. Consequently, they may not always provide accurate identification of potential cases for labeling purposes.
    
\end{itemize}

To address these challenges, we present LeCaRDv2, a Large-Scale Chinese Legal Case Retrieval Dataset. LeCaRDv2 consists of 800 query cases and 55,192 candidate cases selected from a corpus of over 4.3 million Chinese criminal cases. Compared to LeCaRDv1 with only 20 charges in the query set, LeCaRDv2 has three types of query cases covering 50 charges, which can evaluate the effectiveness of retrieval models in the legal domain more comprehensively.
Moreover, with the guidance of official documents published by the Chinese Supreme People's Court, we propose new relevance criteria involving three aspects: characterization, penalty, and procedure. The overall relevance is determined by considering three aspects together. To ensure the quality of the dataset, all annotations are completed by multiple well-trained legal experts who are familiar with all concepts used in the proposed criteria.

Different from previous datasets for legal case retrieval, LeCaRDv2 emphasizes directly retrieving relevant cases from a large legal corpus. 
This is challenging if we only have a limited budget and use simple sampling strategies as those used in LeCaRDv1. To overcome these limitations and label more potential cases with diverse characteristics, we propose a two-level candidate pooling strategy, which includes a retrieval pooling step and a ranking pooling step.
For retrieval pooling, we propose Inverse Provision Frequency (IPF) to measure the similarity of cases based on the law articles. 
Then, to construct the retrieval pool, we employ three distinct methods with different properties: sparse lexical matching, dense semantic retrieval, and the proposed law article similarity. Each method contributes to the construction of a more comprehensive retrieval pool. Moving to the ranking pooling step, we leverage the runs provided by participants in CAIL2021, which are specifically designed for the ranking of LeCaRDv1 and proven to be successful in the context of criminal law cases. These runs are utilized to rank the cases within the retrieval pool, further refining the selection process.

To analyze characteristics of LeCaRDv2, we implement several state-of-the-art models for evaluation. The experimental results indicate that LeCaRDv2 is challenging, and thus more advanced methods for legal case retrieval should be explored. We believe that LeCaRDv2 is a reliable benchmark, and can encourage more fruitful research in the field. It is important to highlight that our dataset provides relevance among case documents. These case documents are openly accessible on the China Judgment Online~\footnote{\url{https://wenshu.court.gov.cn/}}, a website providing collections of case documents published by Chinese government. All sensitive information in cases has been removed or anonymized in advance by the developer of the website, and we have obtained the rightful licences to release all the corresponding case documents and annotations used in LeCaRDv2. 
Users can obtain specific content and potential updates based on the provided case titles.

In summary, LeCaRDv2 is highlighted in the following aspects:

\begin{enumerate} 
\item LeCaRDv2 contains 55,192 candidate cases and 800 query cases covering 50 charges. To the best of our knowledge, LeCaRDv2 is one of the largest Chinese legal case retrieval datasets with the widest coverage of criminal charges. We believe that it can be a reliable benchmark that promotes relevant research in the field.
\item Compared to LeCaRDv1, we design more comprehensive relevance criteria guided by the official documents from the Chinese Supreme People's Court. The new criterion takes into account three aspects, including characterization, penalty, and procedure, providing a more holistic perspective on the relevance of the case.
\item We propose a new two-level candidate pooling strategy to identify potential cases with diverse characteristics. Our strategy consists of a retrieval pooling step and a ranking pooling step. Furthermore, Inverse Provision Frequency (IPF) is proposed to measure the law article similarity of two cases. 
\end{enumerate}

The rest of the paper is organized as follows: Section 2 introduces the related work. In Section 3, the process of dataset construction is elaborated. Then, the experimental setting and results are introduced in Section 4. Finally, Section 5 concludes our work and discusses future work.

\begin{figure*}[t]
\vspace{-3mm}
\centering
\includegraphics[width=0.9\linewidth]{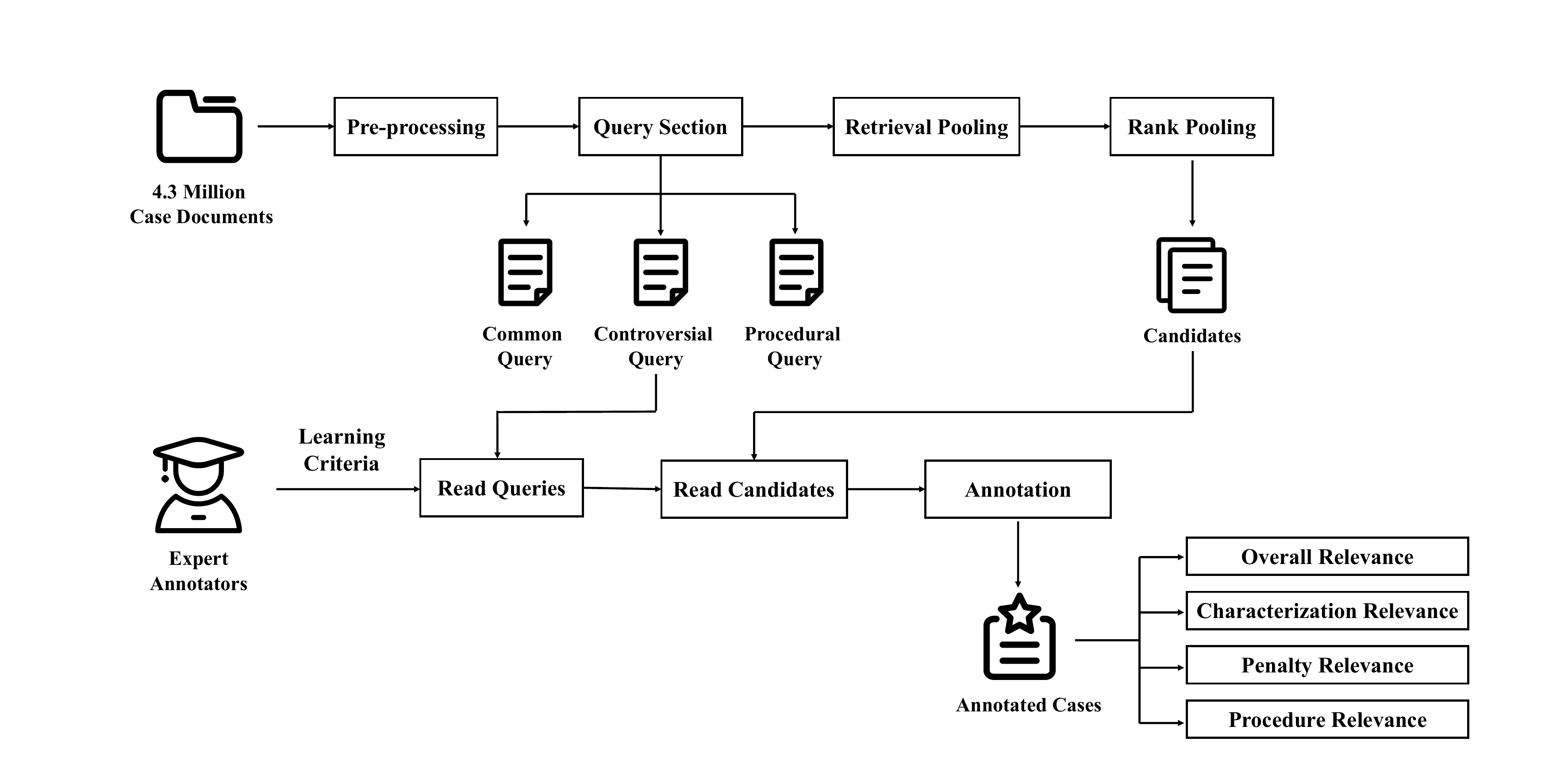}
\vspace{-5mm}
\caption{The data collection and annotation process of LeCaRDv2. }
\label{pipeline}
\vspace{-5mm}
\end{figure*}

\section{Related Work}
We survey related work in terms of datasets, and models of legal case retrieval.

\subsection{Datasets}
Legal case retrieval is an essential and challenging task for legal intelligence systems. Recently, researchers have constructed various benchmarks to promote progress in relevant research. In this section, we provide a summary of existing legal case retrieval benchmarks.

\subsubsection{COLIEE}

As a well-known competition in the legal field, the Competition on Legal Information Extraction/Entailment (COLIEE) aims to achieve state-of-the-art methods of information retrieval using legal texts~\cite{rabelo2022overview,rabelo2021coliee}. Specifically, COLIEE focuses on Canadian case law, which requires reading a query case and identifying relevant support cases from the candidate corpus. For example, COLIEE2020 contains 650 query cases and each query has 200 candidates. Participants need to re-rank a limited number of cases per query. Different from the previous dataset, COLIEE2021 does not provide a candidate pool for each query, which means that participants need to find relevant cases from the entire corpus. It consists of 4,415 case files with 950 query cases, of which 650 queries are for training and 250 queries for testing.

Since the COLIEE dataset belongs to the case law system, its relevance is significantly different from those used in the statutory law systems, e.g., Chinese law system. In COLIEE, the cases cited are considered relevant. Nonetheless, Chinese legal case documents lack such citations. As a result, we must develop new relevance criteria that are suitable for the Chinese legal system.

\subsubsection{CAIL2019-SCM}
To encourage the advancement of the relevant case-matching task, the Chinese AI and Law 2019 Similar Case Matching dataset (CAIL2019-SCM) has been released. CAIL2019-SCM comprises 8,964 triplets, distributed across three legal fields, namely private lending, intellectual property disputes, and maritime affairs. Each triplet contains one query case and two candidate cases, and participants are required to identify which candidate case is more relevant to the query case. However, the task definition of CAIL2019-SCM is substantially different from the actual requirements in practical scenarios, which restricts its application.

\subsubsection{LeCaRDv1}
 LeCaRDv1 is the first legal case retrieval dataset based on the Chinese legal system~\cite{ma2021lecard}. It consists of 107 query cases and 10,700 candidate cases selected from a corpus of over 43,000 Chinese legal case documents. To cover queries with varying difficulties and ranges, LeCaRDv1 introduces a novel query sampling strategy that includes both common queries and controversial queries. In terms of relevance, LeCaRDv1 focuses on the basic facts and proposes four-level relevance criteria based on the critical factor. The critical factors consist of key circumstances and key elements. When two cases share comparable critical factors, they are considered to be related. Inspired by LeCaRDv1, we further refine the relevance criteria and enlarge the size of the dataset. We hope that this enhanced resource will make a more substantial contribution to the development of the legal community.

\subsection{Models}
The development of benchmarks has led to the emergence of models that are tailored for legal case retrieval~\cite{shao2020bert,xiao2021lawformer,bhattacharya2022legal,sailer,li2023thuircoliee,li2023thuircoliee2}. Shao et al.~\cite{shao2020bert} have employed a strategy that involves breaking down legal case texts into multiple paragraphs and utilizing BERT to determine the similarity between these paragraphs, which achieves promising ranking performance. Xiao et al. ~\cite{xiao2021lawformer} introduce a novel attention model and pre-train a Chinese legal language model that can process thousands of tokens. Paheli Bhattacharya et al. ~\cite{bhattacharya2022legal} devise a technique that combines both textual and citation network information to estimate the similarity between legal cases, surpassing the performance of methods that rely solely on one type of information. Large neural language models are data-hungry and obtaining the data for legal case retrieval is expensive. To promote the process of legal case retrieval, it is necessary to develop a large-scale and high-quality dataset.

\section{Dataset Construction}
In this paper, our goal is to construct a large-scale and high-quality dataset for legal case retrieval. The conceptual scheme is illustrated in Figure ~\ref{pipeline}.
In the following section, we first describe the task definition of legal case retrieval. Then, we elaborate on the corpus, queries, and relevance criteria of the dataset. Finally, we describe the human annotation process and the data analysis.

\subsection{Task Defintion}
The task of legal case retrieval is to identify cases related to the query case from the candidate set, which support the decision making process of judges. Specifically, given a query case $q$ and a candidate case set $D = \{d_{1}, d_{2},......, d_{n}\}$ where $n$ is the number of candidate cases, the task is to retrieval top-$k$ related cases $D_q^*=\{d_{i}^* | d_{i}^* \in D)$ with the highest relevance between the query $q$ and the document $d^*_i$.


In the practice of legal case retrieval, queries and candidates are usually long-text documents with complex structures. In general, the query is the fact section of a case document and candidates are entire case documents. Consequently, in this dataset, we construct queries by directly extracting the fact descriptions from the case documents, while omitting other sections such as decision


\subsection{Corpus and Preprocessing}
\begin{figure}[t]
\vspace{-5mm}
\centering
\includegraphics[width=\columnwidth]{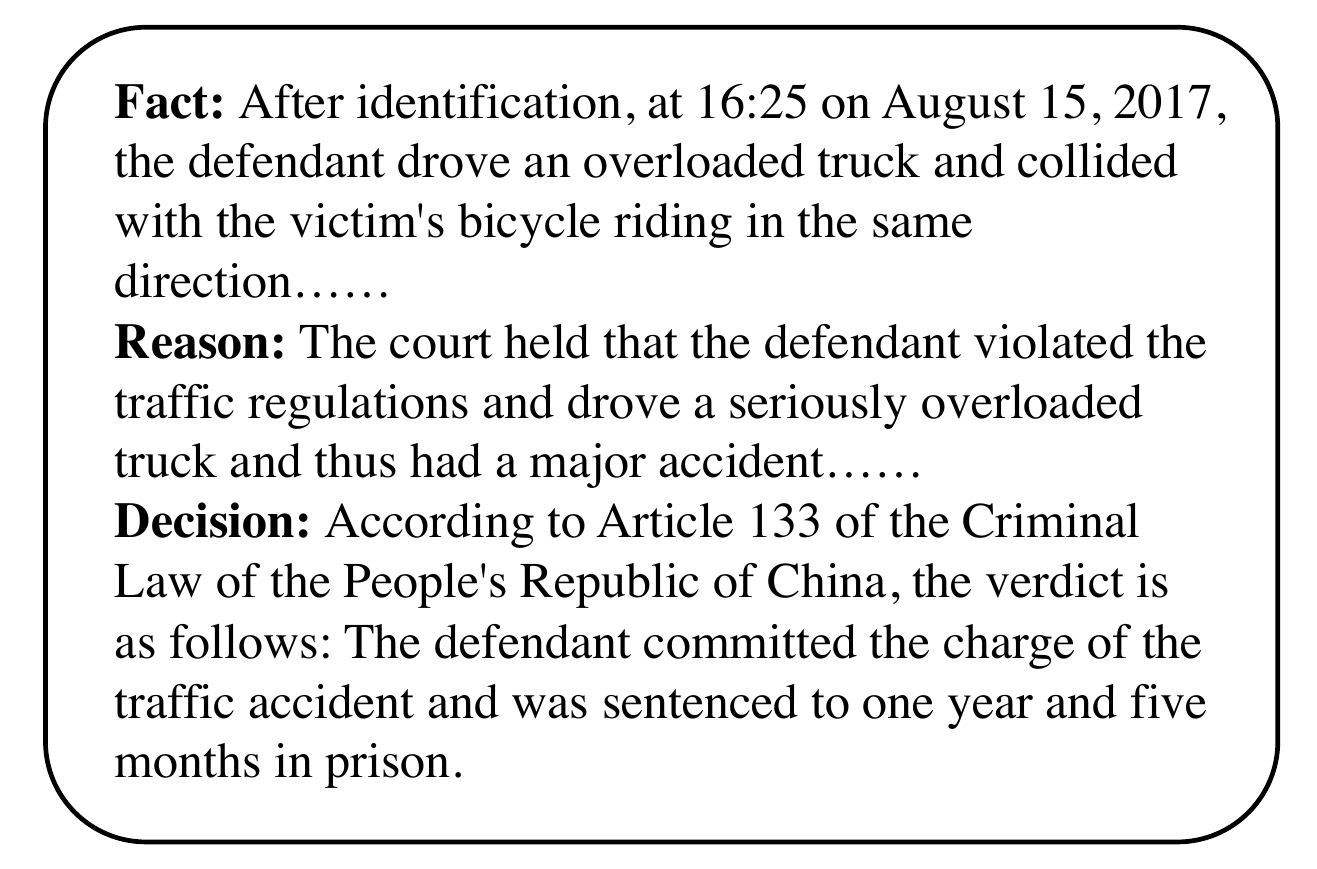}
\caption{An example legal case document.}
\label{case}
\vspace{-5mm}
\end{figure}

To construct this dataset, we collected over 4.3 million criminal case documents from the China Judgment Online. As shown in Figure ~\ref{case}, we divide each case into three sections with regular matching, which consists of fact, reason, and decision. Then, we filter the cases where the fact section is less than 50 characters or involves simple procedures\footnote{Simple procedure refers to the procedure applied to criminal cases with clear facts, simple circumstances, and minor crimes.}.

After data pre-processing, the documents are organized with $(key, value)$ pairs. Moreover, we also extract the charges and the articles of criminal law involved in the case with regular matching. We hope to encourage researchers to explore how to build better retrieval models by utilizing them. It is worth to note that all cases in LeCaRDv2 are publicly available on the China Judgment Online. Users can obtain case details and updates from the website according to the title provided.

\subsection{Query Selection}
Query sampling is essential for the construction of legal case retrieval datasets.
A typical way to construct legal case retrieval queries is to sample cases from the legal corpus randomly and use the fact description section as query text. However, the random sampling of query cases often faces several challenges. On the one hand, the charges divide the cases into different topics, and the number of cases with different charges significantly varies. Random sampling cases to create queries may lead to severe long-tail distribution. On the other hand, users of real legal search systems have different search intents. For example, lawyers and judges may focus more on complicated or controversial queries. The public, which lacks legal knowledge, often wishes to retrieve cases containing as much information as possible. Therefore, it is essential for a high-quality dataset to contain queries with different difficulties and multiple types.

 Inspired by LeCaRDv1~\cite{ma2021lecard}, we apply a sampling strategy that consists of common query, controversial query and procedural query. For common query and controversial query, we expand the coverage of charges used in LeCaRDv1 to include more types of queries. Next, we describe the query sampling strategy in detail.

\subsubsection{Common Query}
The common query refers to cases without second trials and retrials. In other words, legal experts are more likely to come to a consensus in these cases. 

\begin{figure}[t]
\vspace{-5mm}
\centering
\includegraphics[width=\columnwidth]{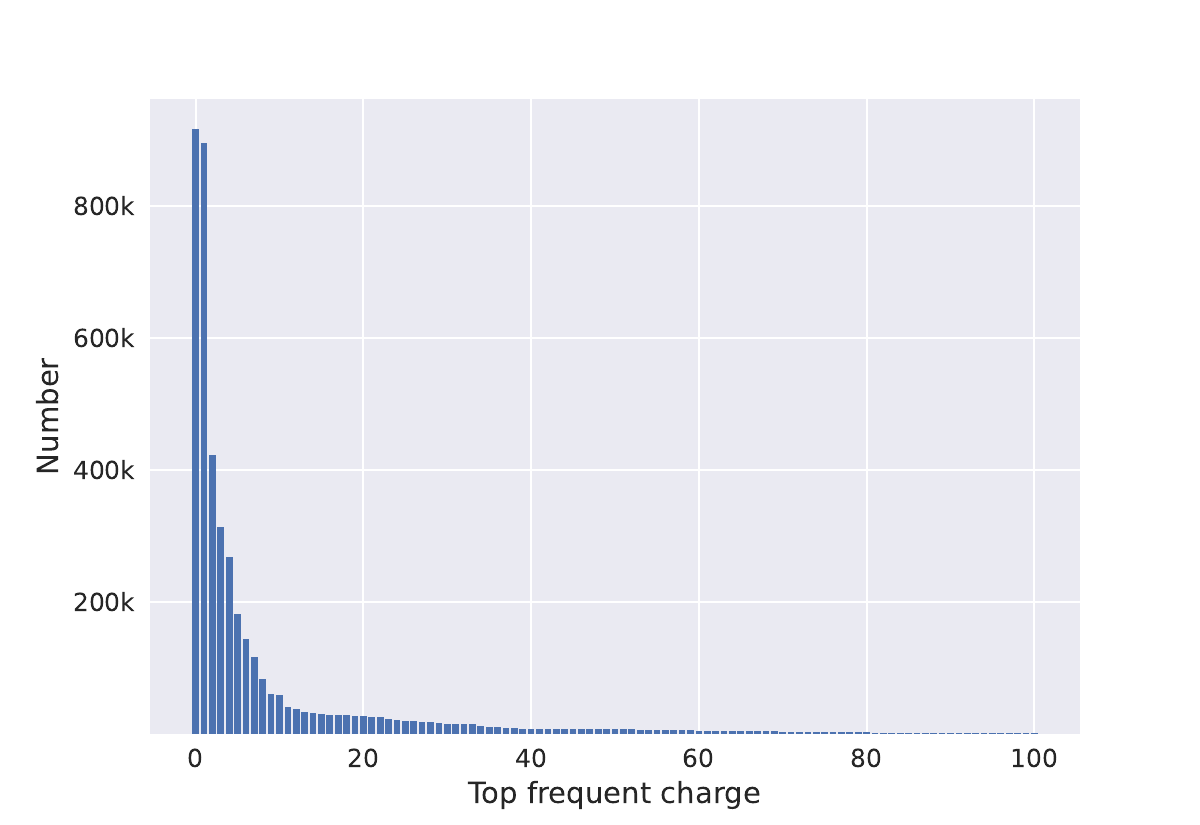}
\vspace{-5mm}
\caption{The distribution of top-100 frequent charges of criminal cases in the last 20 years. X-axis shows the charges IDs sorted by their frequency and Y-axis gives the number of the charges.}
\label{top}
\end{figure}

As shown in Figure ~\ref{top}, we compute the statistics of the distribution of charges of criminal cases in the last 20 years. It can be found that the charges of these cases are severely long-tailed distribution. The number of cases with certain charges can vary from several to hundreds of thousands. Models trained on the dataset with a long-tail distribution may have a strong bias in some charges and cannot accurately estimate the relevance of cases with less frequent charges. To overcome this problem, LeCaRDv1 samples queries from top-20 frequent charges evenly, which account for 86.8\% of all cases. In this paper, we further expand the coverage of charges and select the top-50 frequent charges to construct the common query. 
The top-50 frequent charges account for 96.7\% of the total number of cases, which can cover most of the query cases.

To satisfy the different search intents of the search system users, a high-quality dataset should contain queries with varying levels of difficulty. Following LeCaRDv1, judgment prediction models are employed to sample queries. Specifically, we first train the judgment prediction model following Zhong et al~\cite{zhong2018legal}. Then, we predict the charge of all cases and calculate the prediction entropy to represent the confidence of the prediction model. The prediction entropy can be represented as follows:
\begin{equation}\label{eqn-1} 
  H(c_i) = -\sum_{j=1}^{N}p_{ij} \log p_{ij}
\end{equation}
where $p_{ij}$ denotes the probability that the $ith$ case is predicted to be the $jth$ charge and the $H(c_i)$ is the entropy of the $ith$ case. $N$ represents the total number of charges. For a given case, a higher entropy represents a lower confidence in the prediction of the model. Moreover, the charge with the highest probability is considered the final predicted result of the case. According to prediction correctness and prediction entropy, we classify common queries into the following four categories:
\begin{itemize}[leftmargin=*]
    \item \textbf{true-high entropy}: The model predicts the charge correctly but is not confident. This type of query has moderate difficulty.
    \item \textbf{true-low entropy}: The model predicts the correct charge with high confidence. This type of query is usually easy to solve.
    \item \textbf{false-high entropy}: The model predicts the wrong charge but is uncertain. This type of query is difficult.
    \item \textbf{false-low entropy}: The model predicts the wrong charge with high confidence. This type of query is also difficult.
\end{itemize}

For each charge, we sample three queries in each category. In total, there are $50 \times 12=600$ common queries in our dataset. Compared to LeCaRDv1, the number of common queries is extended seven times. This can provide more training signals and more reliable evaluation results.

\subsubsection{Controversial Query}
The Controversial Query is a case where legal experts have difficulty reaching a consensus. In general, the second trial and retrial cases are more complex and require further discussion. Therefore, we collect all second trial and retrial cases in the corpus to construct the controversial query. Following LeCaRDv1, we sample controversial queries based on the probability of revising judgments. 

Specifically, we calculate the probability of the charges change $P$, the detail of which can be referred to LeCaRDv1~\cite{ma2021lecard}. Then, assuming that the charge $C_{0}$ is changed to $p$ charges in total, we arrange them in descending order $P_{C_0 \rightarrow C_1}, P_{C_0 \rightarrow C_2}, ..., P_{C_0 \rightarrow C_p}$.  $Top-q$ frequent charges are selected and $q$ satisfies the following conditions:

\begin{equation}\label{eqn-2} 
  \sum_{i=1}^{q} P_{C_0 \rightarrow C_i} \geq 0.5
\end{equation}



   



The controversial queries are sampled from the selected charges. There are a total of 100 controversial queries in this dataset.

\subsubsection{Procedural Query}
The procedural query refers to cases where procedural legality is disputed in criminal proceedings. An intuitive approach is to select cases related to procedural law. Specifically, substantive law concerns the set of legal principles that govern the behavior of individuals and society generally. Procedural law, also known as adjective law, refers to the regulations and guidelines that govern the processes involved in creating, implementing, and enforcing substantive law. The criteria for relevance under procedural law may vary from those used in substantive law.

To formulate the procedural query, we gather the relevant keywords\footnote{These keywords are available in GitHub https://github.com/THUIR/LeCaRDv2.} related to procedural disputes and utilize them to identify a set of appropriate procedural queries. Then, for the top-50 frequent charges, we sample two cases for each charge based on the keyword filtering above as the procedural queries. Finally, a total of 100 procedural cases are collected in this section.

\subsection{Candidate Pooling}
In practice, it is impractical and expensive to annotate the relevance of a query to all other cases. To allocate limit judging resources to more promising cases, depth-$k$ pooling has been applied in many retrieval datasets~\cite{rabelo2021coliee,arora2018challenges,ma2021lecard,clarke2009overview,nguyen2016ms}. 
This approach involves obtaining a document pool from existing retrieval models and then having annotators label their relevance. For instance, LeCaRDv1 adopts three lexical matching models for pooling and merge the top-100 retrieved cases into the final pool. However, this simple pooling strategy retrieves cases with similar properties, which may does not fully exploit promising cases. 

To address this issue, we propose a two-level pooling strategy comprising a retrieval pooling step and a rank pooling step. The retrieval pooling step aims to form the candidate set with diversity by combining sparse lexical matching, dense semantic retrieval, and law article similarity. Subsequently, the ranking pooling further prioritizes the cases in the retrieval pool using runs submitted by previous participants in CAIL2021, with the goal of identifying the most promising cases for annotation.

\subsubsection{Retrieval Pooling}
For each query, the retrieval pooling step selects 100 candidate cases from the large corpus. It combines sparse lexical matching, dense semantic retrieval, and law article similarity to improve the diversity of candidate sets.

\textbf{Sparse lexical matching}: Sparse lexical matching methods calculate the similarity based on the same words between the query and the candidate. In this section, we employ BM25, a classical lexical matching method, to retrieve the relevant case documents. To be specific, we first apply jieba~\footnote{\url{https://github.com/fxsjy/jieba}} to split the Chinese sentences into words. Then we remove the stop words and retrieve the top-100 cases for each query from the entire corpus.

\textbf{Dense semantic retrieval}: In recent years, with the development of pre-trained language models, semantic-based dense retrieval models have attracted considerable attention. Dense retrieval models typically employ complex neural networks to encode query and document as $h_q$ and $h_d$ respectively. The semantic relevance scores are then calculated by applying the dot product or cosine similarity to their encoded representations. In general, dense retrieval models can better capture the semantic information of the context through complex interactions. To construct the pool of LeCaRDv2 with dense retrieval models, we apply RoBERTa~\cite{liu2019roberta} as the backbone and pre-train it on a legal corpus with whole word mask (WWM) task. The top-100 relevant case documents for each query are retrieved with this model.

\textbf{Law article similarity}: The law article similarity aims to calculate the relevant score by the law article cited in the cases. Since the definition of relevance in the legal field differs from that of the general domain, it is not sufficient to only use text-based similarity methods, i.e., lexical matching and semantic retrieval. Therefore, inspired by inverse document frequency (IDF)~\cite{TF-IDF}, we propose Inverse Provision Frequency (IPF) to measure the similarity of cases in terms of the law articles. To be specific, we extract the criminal law articles involved in each case document. Given a specific law article $P_{i}$, the IPF value is as follows:
\begin{equation}\label{eqn-3} 
  IPF_{P_{i}} = \log \dfrac {|D|} {freq(P_{i},D)}
\end{equation}

where $|D|$ is the size of the corpus and $freq(P_{i},D)$ represents the number of cases containing $P_i$ in corpus $D$. The design of the IPF is based on the intuition that law articles cited in a large number of cases may not contain information that is important for a particular query. In other words, it cannot provide enough information to determine relevance. The law article similarity of the two cases is calculated by summing the IPF values of all co-cited law articles. As described above, we use IPF to retrieve the top-100 relevant case documents for each query.

To merge the cases retrieved by the three methods described above into a single pool, we divide them into four groups:
\begin{itemize}[leftmargin=*]
    \item \textbf{Group 1}: top-25 cases retrieved retrieved by any of the above. Note that the total number of cases in this group may be less than 75, since there may be duplicate cases retrieved with different methods.

    \item \textbf{Group 2}: After filtering out the cases in group 1, the remaining cases occurred in the top-100 cases in all three retrieval models.

    \item \textbf{Group 3}: After filtering out the cases in group 1, the remaining cases occurred in the top-100 cases for exactly two retrieval models.
    
    \item \textbf{Group 4}: After filtering out the cases in group 1, the remaining cases occurred in only one retrieval model.
    
\end{itemize}

For each query, the candidate pool contains all cases in group 1. Then, we select cases from group 2 to supplement the candidate pool to 100 cases. If the total number of cases in group 1 and group 2 is less than 100, the leftover cases are sampled from group 3. Similarly, cases from group 4 will be added to the candidate pool if group 3 does not have enough cases. In short, the retrieval pool contains cases with the highest scores in each retrieval model and cases retrieved by multiple models. Given a query, we form a pool of candidate cases with a depth of 100.

\subsubsection{Rank Pooling}
After retrieval pooling, we conduct rank pooling with the runs provided by participants in the CAIL2021 legal case retrieval track (CAIL2021-LCR). CAIL2021-LCR provides a candidate case pool with a depth of 100 for each query, and participants need to re-rank a limited number of cases per query. These runs are all designed for ranking Chinese criminal law cases, which is consistent with our dataset. To be specific, we collect three award-winning runs in CAIL2021-LCR, which are employed to rank the retrieval case pool. We preserve the top-30 cases of each run and divide them into 4 four groups.

\begin{itemize}[leftmargin=*]
    \item \textbf{Group 1}: top-5 cases ranked with the above three runs. There may be less than 15 cases in this group since some cases may be duplicates.
    \item \textbf{Group 2}: After filtering out the cases in group 1, the remaining cases occurred in the top-30 cases in all three re-rank runs.
    \item \textbf{Group 3}: After filtering out the cases in group 1, the remaining cases occurred in the top-30 cases for exactly two re-rank runs.
    \item \textbf{Group 4}: After filtering out the cases in group 1, the remaining cases occurred in only one re-reank run.
\end{itemize}

The rank pool contains all the cases in group 1 and replenishes them to 30 in the order of group 2, group 3 and group 4 priority. The rank pool contains cases with the highest scores in each run and ranked in top-30 by multiple runs. Thus we have generated the rank case pool with a depth of 30 for each query, where cases are considered to be more potentially relevant. The rank pool will be available for legal experts to annotate.

\subsection{Relevance judgment Criteria}
The Supreme People's Court of China has published a guidance document~\footnote{\url{http://www.hncourt.gov.cn/public/detail.php?id=181775}} for case relevance under the Chinese legal system. There are three main aspects of relevant cases: basic facts, focus of the disputes, and application of law.

The relevance criteria of LeCaRDv1 focus on basic facts while ignoring the focus of the disputes and application of law. For this dataset, we design a more comprehensive relevance criteria, which follow the official guidance better. Specifically, the advancement of our relevance criteria is reflected in two aspects. First, the annotator needs to judge the Overall Relevance, which follows the official documents strictly. Second, we propose unique relevance evaluation criteria in three aspects: characterization, penalty, and procedure, which comprehensively cover the kinds of information needs of users in judicial practice to conduct legal case retrieval. Next, we describe their definitions in detail.

\textbf{Characterization Relevance}: Characterization Relevance focuses on the basic fact of the case. As with LeCaRDv1, critical factor is employed to measure Characterization Relevance. Critical factor has a substantial impact on the trial of the case. Before annotation, the assessors need to determine whether the query case constitutes a crime and what crime it constitutes. The Characterization Relevance is defined as:

\emph{Two cases are defined as relevant in Characterization if the similarity between their critical factors is high.}

The critical factors consist of key circumstances and key constitutive elements of the crime (key elements). Key elements are the legal concept abstraction of key circumstances. The different legal elements can lead to different judgments. The Characterization Relevance labeling is in a four-level setting ranging from 1 to 4 with increasing relevance. Detailed descriptions of the relevant labels are as follows:

\begin{itemize}[leftmargin=*]
    \item \textbf{Label-1}: Both key elements and key circumstances are irrelevant.
    \item \textbf{Label-2}: Key circumstances are relevant but key elements are irrelevant.
    \item \textbf{Label-3}: Key elements are relevant but key circumstances are irrelevant.
    \item \textbf{Label-4}: Both key elements and key circumstances are relevant. 
\end{itemize}

\begin{figure*}[t]
	\centering
	
	\subfigure[Beginning Screen ]{
		\begin{minipage}[t]{0.31\linewidth}
			\centering
                \label{begining}
			\includegraphics[width=1.0\linewidth]{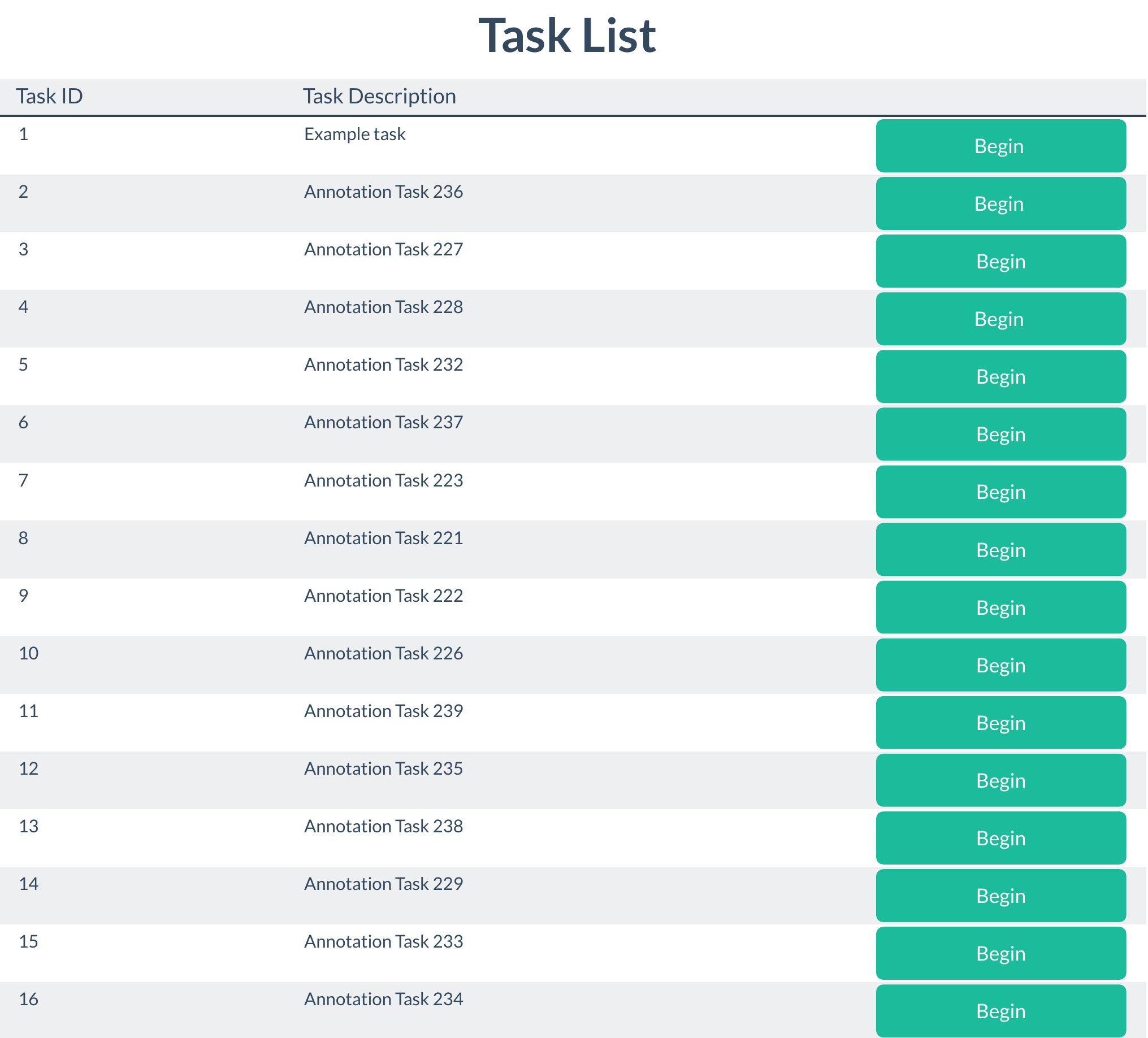}
                
		\end{minipage}
	}
	\subfigure[Query Display]{
		\begin{minipage}[t]{0.31\linewidth}
                \label{query}
			\centering
			\includegraphics[width=1.0\linewidth]{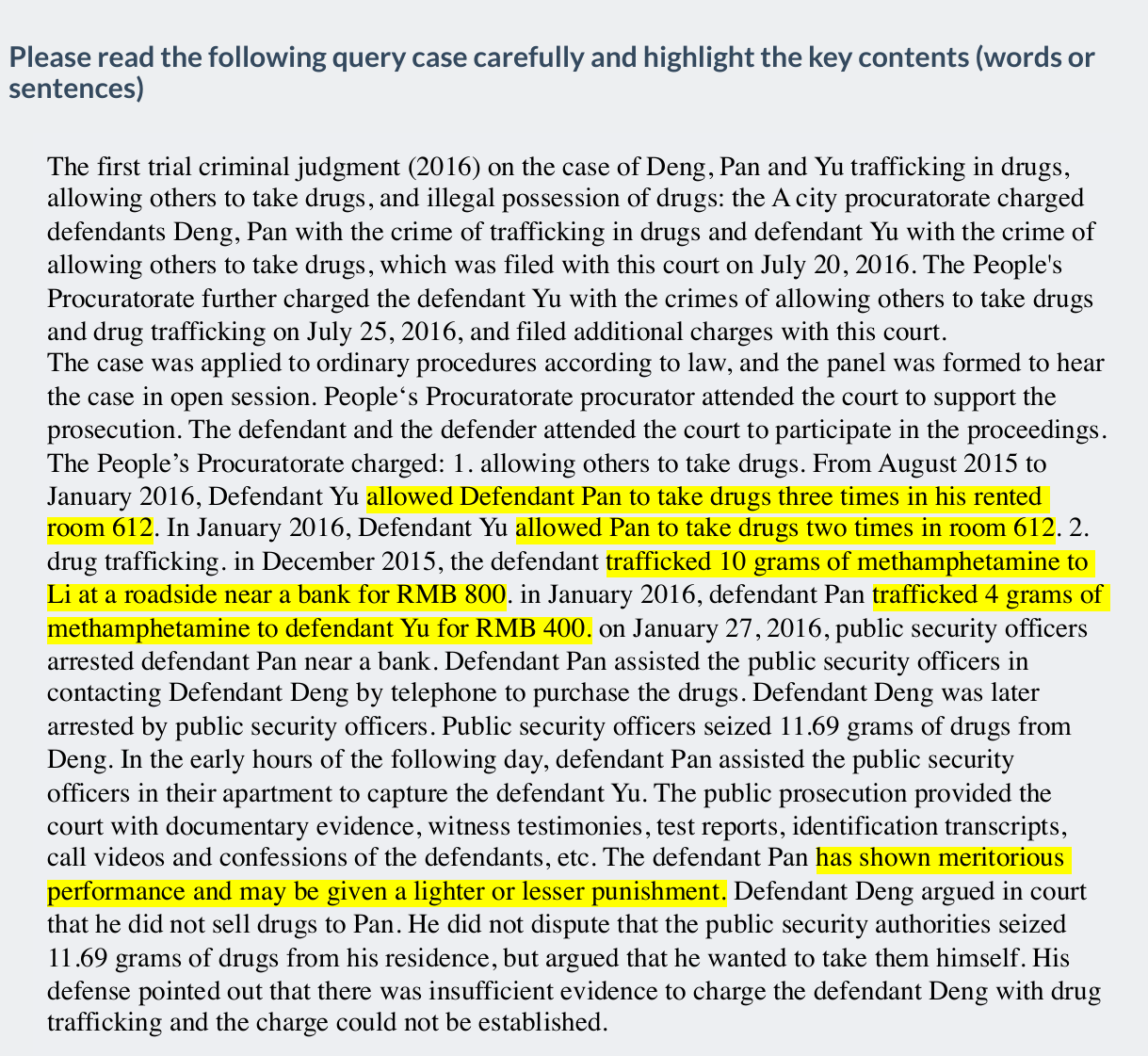}
		\end{minipage}
	}
 	\subfigure[Candidate Display]{
		\begin{minipage}[t]{0.31\linewidth}
			\centering
                \label{candidate}
			\includegraphics[width=1.0\linewidth]{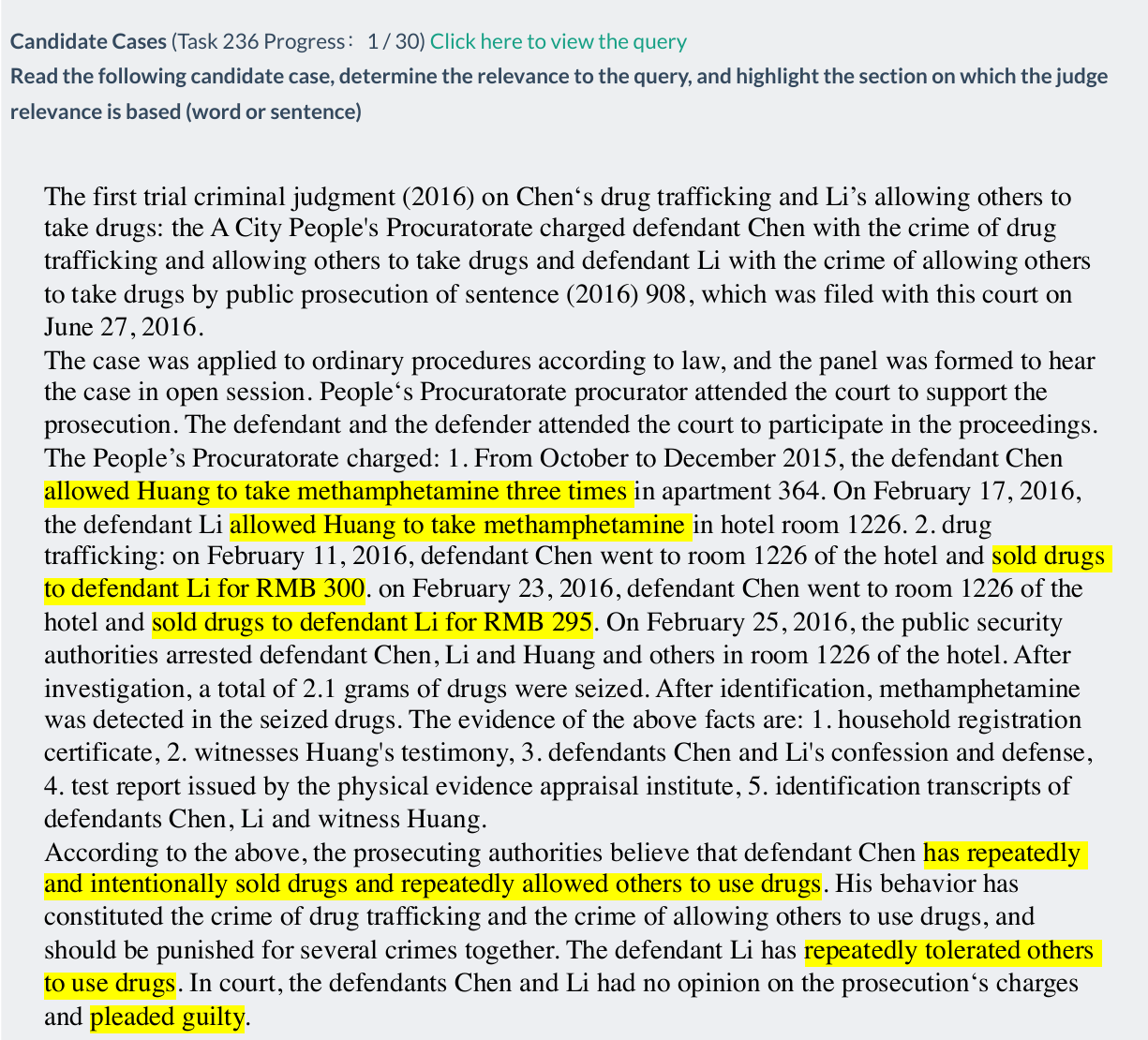}
		\end{minipage}
	}
 \vspace{-5mm}
	\caption{An example of the THUIR search experiment platform. The annotators are required to highlight key content that affects the judgment of relevance.}
	\label{platform}
\vspace{-5mm}
\end{figure*}

\textbf{Penalty Relevance}: Penalty Relevance focuses on the circumstances of sentencing. In reality, we note that although the basic facts of two cases are similar, the sentences may be different. Before annotation, the annotator needs to determine the subjective factors of the defendant and what the defendant has committed. The Penalty Relevance is defined as:

\emph{Two cases are defined as relevant in Penalty if the similarity between their circumstances of sentencing is high.}

The circumstances of sentencing consist of the commission of the crime and the individual circumstances of the offender. The commission of the crime includes the criminal pattern, specific means, tools, harmful results and the impact on society of criminal behavior. The individual circumstances of the offender include the offender's age, experience (previous convictions, repeat offenses), attitude after the crime (surrender, covering victims' losses), etc. The Penalty Relevance labeling is also in a four-level setting ranging from 1 to 4 with increasing relevance. Detailed descriptions of the relevant labels are as follows:

\begin{itemize}[leftmargin=*]
    \item \textbf{Label-1}: Both commissions of crime and offender circumstances are irrelevant.
    \item \textbf{Label-2}: Offender circumstances are relevant but commissions of crime are irrelevant.
    \item \textbf{Label-3}: Commissions of crime are relevant but offender circumstances are irrelevant.
    \item \textbf{Label-4}: Both commissions of crime and offender circumstances are relevant.
\end{itemize}

\textbf{Procedure Relevance}
Procedure Relevance concerns the procedural controversy of the case. Procedural controversy refers to the dispute about the legality of procedural facts in criminal proceedings, which directly affects the application of the law~\cite{shao2023intent}. The Procedure Relevance is defined as:

\emph{Two cases are defined as relevant in Procedure if the similarity between their procedural controversy is high.}

The procedural controversy includes procedural issues and procedural facts. 
Procedural issues are disputes involving procedural law, such as the exclusion of illegal evidence, and judicial jurisdiction. Procedural facts are the specific circumstances of the procedural dispute. The Procedure Relevance labeling is in a four-level setting ranging from 1 to 4 with increasing relevance. 
\begin{itemize}[leftmargin=*]
    \item \textbf{Label-1}: Procedural issues and procedural facts are irrelevant.
    \item \textbf{Label-2}: Procedural facts are relevant but procedural issues are irrelevant.
    \item \textbf{Label-3}: Procedural issues are relevant but procedural facts are irrelevant.
    \item \textbf{Label-4}: Both procedural issues and procedural facts are relevant.
\end{itemize}

For Overall Relevance, annotators are required to follow the official guidance criterion and take the characterization, penalty, and procedure relevance as a whole into consideration. We provide two guide cases with relevant score explanations to annotators for better understanding. The Overall Relevance labeling is in a four-level setting which consists of Irrelevant, Somewhat irrelevant, Fairly relevant, and Highly relevant. It is worth noting that there is no explicit mapping function between the  Overall Relevance and the sub-relevance. The annotators have discretionary authority and makes judgments based on their knowledge.

\begin{table*}[t]
\vspace{-5mm}
\caption{The statistics of legal class case retrieval datasets.}
\label{statistics}
\begin{tabular}{@{}ccccccc@{}}
\toprule
DATASETS                      & LeCaRDv1  & CAIL2019-SCM & CAIL2022-LCR & COLIEE2020 & COLIEE2021 & LeCaRDv2 \\ \midrule
Language                      & Chinese & Chinese      & Chinese      & English    & English    & Chinese  \\
\#Queries                     & 107     & 8264         & 130          & 650        & 900        & 800      \\
\# Candidate cases/query      & 100     & 2            & 100          & 200        & 4,415      & 55,192   \\
Avg.length per case document  & 8,275   & 676          & 2,707        & 3,232      & 1,274      & 4,766    \\
\#Avg.relevant case per query & 10.33   & 1            & 11.53        & 5.15       & 4.73       & 20.89    \\ \bottomrule
\end{tabular}
\vspace{-5mm}
\end{table*}

\subsection{Human Annotation}
Our relevance annotators consist of 41 legal experts majoring in criminal law, who have all passed the National Uniform Legal Profession Qualification Examination and are familiar with the cases in our dataset. To ensure the quality of annotation, all annotators first go through several hours of interpretation to ensure a sound understanding of the concept in the criteria. Then, we verify the quality of annotators with several example tasks. The creator of the criteria, who holds a Ph.D. in criminal law, makes corrections according to their annotation in example cases to ensure consistent understanding among all annotators. Only annotators who have passed the training are allowed to perform official annotation. Each annotation task is completed repeatedly by three different annotators. We measure the annotation quality with Kappa~\cite{cohen1960coefficient,fleiss1973equivalence}. The Kappa value for the Overall Relevance is 0.5190, which indicates that LeCaRDv2 is a high-quality manual labeled dataset. The average of the three annotation results is regarded as the final relevance score. For each annotation task, we pay the legal expert 6.60 dollars.

For efficient annotation, we build an annotation platform to help annotators browse cases and make judgments about their relevance. Figure ~\ref{platform} shows an example of the annotation platform. 
Figure ~\ref{begining} is the beginning screen of the platform, from which the annotator can select the annotation task.
Each annotation task contains one query case and thirty candidate cases. The candidate cases are presented in a randomized manner to remove the rank bias of the assessors. In other words, it prevents the annotators from over-valuing the higher-ranked cases and under-valuing the later ones.
Figure ~\ref{query} and \ref{candidate} are the query and candidate display screens respectively. 
During the annotation process, the annotators are also required to highlight key content that affects the judgment of relevance. After careful reading, the annotator needs to make a judgment about four relevance scores between the candidate cases and the query. For Characterization, Penalty, and Procedure relevance, we provide the option with a label=0, meaning that no legal issue is involved between cases in that aspect. 


\subsection{Data Analysis}
In this section, we analyze the different attributes of LeCaRDv2.

\subsubsection{Data Size}
Detailed statistics of LeCaRDv2 and other popular legal case retrieval datasets are shown in Table ~\ref{statistics}. There are some datasets i.e. LeCaRDv1, CAIL2019-SCM, CAIL2022-LCR, and COLIEE2020 providing a limited number of candidate cases for each query. While COLIEE2021 and LeCaRDv2 require determining relevant documents from the entire corpus, which makes the task more difficult. Compared with LeCaRDv1, LeCaRDv2 is closer to the real-world retrieval task, which requires the model with both efficiency and effectiveness.

From the statistics, we can find that LeCaRDv2 is the largest Chinese legal case retrieval dataset with tens of thousands of annotated data. Benefiting from the candidate case pooling strategy, more promising cases are annotated, which further increases the number of relevant cases.

\subsubsection{Data Distribution}
\begin{figure}[t]
\centering
\includegraphics[width=\columnwidth]{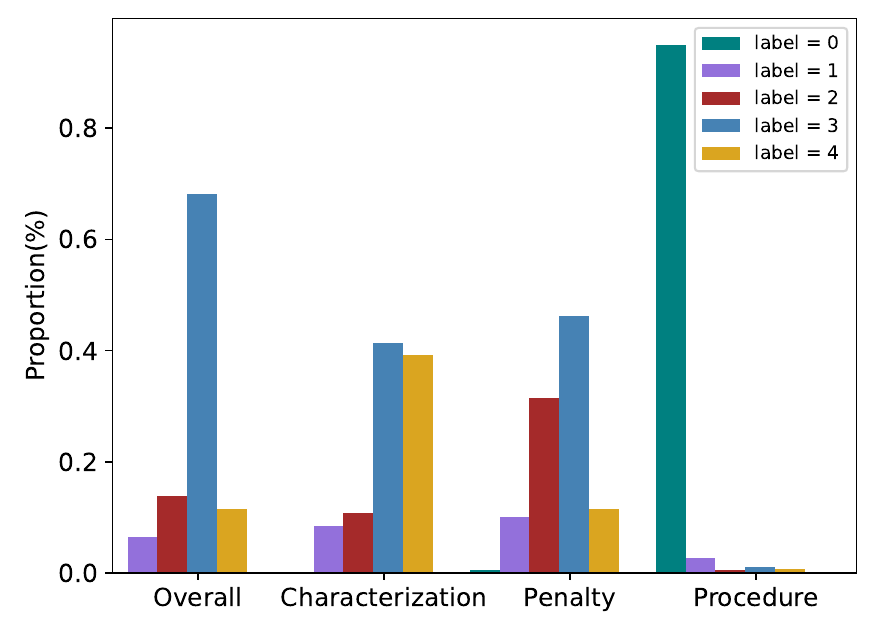}
\vspace{-5mm}
\caption{Distribution of cases with different relevance labels. Label=0 means that the query and candidate do not involve this type of relevance.}
\label{distribution}
\vspace{-5mm}
\end{figure}

Figure ~\ref{distribution} shows the distribution of query-candidate pairs with different relevance levels.  We can find that most annotated cases are fairly relevant in Overall Relevance, We think it benefits from the large corpus, which can provide sufficient relevant cases, and the candidate pooling strategy, which can determine the more promising relevant cases to annotate. For Characterization Relevance, the number of fairly relevant and highly relevant cases is almost equal. Moreover, it is worth noting that there are some cases that do not involve Penalty Relevance and Procedure Relevance judgment. For Procedure Relevance, since the majority of cases do not involve procedural disputes in China, there are numerous cases with a procedural relevance label of 0. Still, Procedure Relevance is an important component of the relevance in legal cases and deserves further study.

\begin{table*}[t]
\caption{Zero-shot and finetune performance of various baselines on LeCaRDv2. Under zero-shot setting, all queries are applied for evaluation. Under fine-tuning setting, there are 160 queries for testing. The best method in each column is marked in bold.}
\label{result}
\begin{tabular}{lcccccccc}
\hline
\multicolumn{1}{l|}{\multirow{2}{*}{Models}} & \multicolumn{4}{c|}{Zero-shot}                                                             & \multicolumn{4}{c}{Fine-tune}                                         \\ \cline{2-9} 
\multicolumn{1}{l|}{}                        & R@100           & R@200           & R@500           & \multicolumn{1}{c|}{R@1000}          & R@100           & R@200           & R@500           & R@1000          \\ \hline \hline
\multicolumn{9}{c}{\textbf{Traditional Retrieval Models}}                                                                                                                                                         \\ \hline
\multicolumn{1}{l|}{BM25}                    & \textbf{0.6262} & \textbf{0.6629} & 0.6946          & \multicolumn{1}{c|}{0.7207}          & \textbf{0.6050} & \textbf{0.6428} & 0.6735          & 0.7015          \\
\multicolumn{1}{l|}{QLD}                     & 0.5984          & 0.6576          & \textbf{0.7065} & \multicolumn{1}{c|}{\textbf{0.7424}} & 0.5749          & 0.6354          & 0.6882          & 0.7222          \\ \hline \hline
\multicolumn{9}{c}{\textbf{Generic Pre-trained Models}}                                                                                                                                                           \\ \hline
\multicolumn{1}{l|}{Chinese-BERT-WWM}        & 0.1165          & 0.1526          & 0.2184          & \multicolumn{1}{c|}{0.2805}          & 0.3849          & 0.5026          & 0.6649          & 0.7797          \\
\multicolumn{1}{l|}{Chinese-RoBERTa-WWM}     & 0.3753          & 0.4739          & 0.6152          & \multicolumn{1}{c|}{0.7126}          & 0.4136          & 0.5330           & 0.6964          & 0.7998          \\
\multicolumn{1}{l|}{Bert\_xs}                & 0.0453          & 0.0614          & 0.0949          & \multicolumn{1}{c|}{0.1343}          & 0.2074          & 0.2750           & 0.3935          & 0.4941          \\
\multicolumn{1}{l|}{Lawformer}               & 0.2432          & 0.3040          & 0.4054          & \multicolumn{1}{c|}{0.4833}          & 0.3651          & 0.4851          & 0.6443          & 0.7629          \\ \hline \hline
\multicolumn{9}{c}{\textbf{Retrieval-oriented Pre-trained Models}}                                                                                                                                                \\ \hline
\multicolumn{1}{l|}{Condenser}               & 0.2215          & 0.2987          & 0.4321          & \multicolumn{1}{c|}{0.5452}          & 0.3982          & 0.5003          & 0.6761          & 0.7969          \\
\multicolumn{1}{l|}{coCondenser}             & 0.2255          & 0.3093          & 0.4460           & \multicolumn{1}{c|}{0.5514}          & 0.3998          & 0.5024          & 0.6861          & 0.8036          \\
\multicolumn{1}{l|}{SEED}                    & 0.3544          & 0.4474          & 0.5745          & \multicolumn{1}{c|}{0.6657}          & 0.4201          & 0.5437          & \textbf{0.7160} & 0.8132          \\
\multicolumn{1}{l|}{RetroMAE}                & 0.3193          & 0.3947          & 0.5010           & \multicolumn{1}{c|}{0.5821}          & 0.4210          & 0.5397          & 0.7093          & \textbf{0.8174} \\ \hline
\end{tabular}
\vspace{-5mm}
\end{table*}

\section{Experiment}
In this section, we implement some state-of-the-art models to evaluate LeCaRDv2. We first introduce the benchmark settings and baselines. Then, we report the experimental results and perform a detailed analysis.

\subsection{Benchmark Settings}
We conduct experiments on state-of-the-art models with zero-shot and fine-tuning settings. Under the zero-shot setting, the model is not trained with any annotated data, which is suitable for legal systems that lack training data. All annotated data in LeCaRDv2 are employed to evaluate.
Under the fine-tuning setting, we sampled 20\% cases from each charge as the test set. There are 640 queries for training and 160 queries for testing. Since we focus on retrieval performance in large corpus, we adopt recall as the evaluation metric.

\subsection{Baselines}
We adopt three types of widely-used retrieval models as baselines, including Traditional Retrieval Models, Generic Pre-trained Models, and Retrieval-oriented Pre-trained Models.
For Pre-trained Models, the dual encoder architecture is applied to retrieve relevant cases from the entire corpus. 
\begin{itemize}[leftmargin=*]
    \item \textbf{Traditional Retrieval Models}
    
    \begin{itemize}
        \item[-] \textbf{BM25}~\cite{robertson2009probabilistic} is a robust traditional retrieval model based on word matching.
        \item[-] \textbf{LMIR}~\cite{robertson2009probabilistic} is a highly effective strong baseline mode based on Dirichlet smoothing.
        
    \end{itemize}  

    \item \textbf{Generic Pre-trained Models} 
    
   \begin{itemize}
        \item[-] \textbf{Chinese-BERT-WWM}~\cite{chinese-bert-wwm} is a multi-layer transformer trained with Whole Word Mask (WWM) and Next Sentence Prediction (NSP) tasks.
        \item[-] \textbf{Chinese-RoBERTa-WWM}~\cite{chinese-bert-wwm} has the same architecture as Chinese-Bert-WWM, which is trained in enlarged datasets with only WWM task. 
        \item[-] \textbf{BERT\_xs} ~\footnote{\url{http://zoo.thunlp.org/xs-bert}} is the Bert specialized in criminal law, which is trained with several million Chinese case documents.
        \item[-] \textbf{Lawformer}~\cite{xiao2021lawformer} aims to process long legal cases, which employ Longformer~\cite{beltagy2020longformer} as the backbone.
    \end{itemize}
    
    \item \textbf{Retrieval-oriented Pre-trained Models} 
    
    \begin{itemize}
        \item[-] \textbf{Condenser}~\cite{gao2021condenser} designs the skip connection to force information to be integrated into the [CLS] token.
        \item[-] \textbf{coCondenser}~\cite{gao2021unsupervised} utilizes unsupervised contrastive learning to warm up the vector space based on Condenser.
        \item[-] \textbf{SEED}~\cite{lu2021less} applies a weak decoder to enhance the training of the encoder, which achieves state-of-the-art performance in ad-hoc retrieval tasks.
        \item[-] \textbf{RetroMAE}~\cite{liu2022retromae} designs a harder decoding process to achieve better retrieval performance.
    \end{itemize}
\end{itemize}

We use the pyserini toolkit~\footnote{\url{https://github.com/castorini/pyserini}} to implement BM25, which with default parameters. For generic pre-trained models, we directly load their checkpoints in huggingface~\footnote{\url{https://huggingface.co/models}}. For retrieval-oriented pre-trained models, we reproduced their work on the legal corpus with their open-source code as there are no available Chinese versions of them. We adopt BERT to initialize retrieval-oriented pre-trained models. Following the previous work~\cite{zhan2020repbert,karpukhin2020dense}, negative samples are BM25 negatives and the ratio of positives and negatives is 1:32.

\subsection{Experimental Results}

The performance of baselines on LeCaRDv2 is shown in Table ~\ref{result}. In the zero-shot setting, we directly measure the performance of all queries with the pre-trained language model. In the fine-tuning setting, we use 640 queries for training and 160 queries for testing. From the experimental results, we can derive the following observation:

\begin{itemize}[leftmargin=*]
    \item Both in the zero-shot and fine-tuning settings, the traditional retrieval methods show competitive performance on legal case retrieval task.
    \item Under the Zero-shot setting, generic pre-trained models generally perform worse than traditional retrieval models. Despite training with extensive criminal law data, the performance of BERT\_xs is worse than that of BERT. We guess that this is because the pre-training objectives of BERT\_xs damage the robustness of the [CLS] embedding, which is not suitable for dense retrieval. Correspondingly, Chinese-RoBERTa-WWM achieves surprising results. This indicates that the next sentence prediction task may not be helpful for dense retrieval.
    \item Retrieval-oriented pre-training models generally have better performance than generic pre-training models. This indicates that retrieval-oriented pre-training tasks rather than general NLP tasks are more helpful for retrieval. 
    \item With the guidance of labeled data, the performance of pre-trained models is further improved. However, in some metrics i.e. R@100, R@200, BM25 achieves the best performance, which encourages the community to propose more pre-trained language models for legal case retrieval.
    \item In short, LeCaRDv2 is a challenging retrieval task. Existing pre-trained language models perform worse on legal documents than in the general domain due to length restrictions and different relevance definitions. We are confident that with more test data, LeCaRDv2 can better test the significant effectiveness of the proposed models. It is worth investigating in the future to design better retrieval models in the legal domain.
\end{itemize}

\section{Conclusion}

In this paper, we release LeCaRDv2 as a new and challenging dataset for legal case retrieval. LeCaRDv2 consists of 800 queries covering 50 charges and 55,192 candidate cases, which is one of the largest Chinese legal case retrieval datasets with the widest coverage of criminal charges. We enrich criteria of legal relevance based on LeCaRDv1, which covers characterization, penalty, procedure three aspects and Overall Relevance. Moreover, we propose a novel candidate pooling strategy to identify potential cases with diverse characteristics. We evaluated several competitive baselines on LeCaRDv2. The experimental results show that LeCaRDv2 is a challenging retrieval task and further efforts are needed to promote the development of legal case retrieval. In the future, we will continue to expand the size of this dataset and start focusing on civil law cases. Moreover, we will attempt to extract the highlights of legal experts in determining the relevance of cases for contributing to the community.

\clearpage
\balance
\bibliographystyle{ACM-Reference-Format}
\bibliography{sample-base.bib}
\end{document}